# COMPUTATIONAL POWER AND THE SOCIAL IMPACT OF ARTIFICIAL INTELLIGENCE

Tim Hwang[1]

*Machine learning is a computational process. To that end, it is inextricably tied to computational power - the tangible material of chips and semiconductors that the algorithms of machine intelligence operate on. Most obviously, computational power and computing architectures shape the speed of training and inference in machine learning, and therefore influence the rate of progress in the technology. But, these relationships are more nuanced than that: hardware shapes the methods used by researchers and engineers in the design and development of machine learning models. Characteristics such as the power consumption of chips also define where and how machine learning can be used in the real world.*

*In a broader perspective, computational power is also important because of its specific geographies. Semiconductors are designed, fabricated, and deployed through a complex international supply chain. Market structure and competition among companies in this space influence the progress of machine learning. Moreover, since these supply chains are also considered significant from a national security perspective, hardware becomes an arena in which government industrial and trade policy has a direct impact on the fundamental machinery necessary for artificial intelligence (AI).*

*This paper aims to dig more deeply into the relationship between computational power and the development of machine learning. Specifically, it examines how changes in computing architectures, machine learning methodologies, and supply chains might influence the future of AI. In doing so, it seeks to trace a set of specific relationships between this underlying hardware layer and the broader social impacts and risks around AI. On one hand, this examination shines a spotlight on how hardware works to exacerbate a range of concerns around ubiquitous surveillance, technological unemployment, and geopolitical conflict. On the other, it also highlights the potentially significant role that shaping the development of computing power might play in addressing these concerns.*

\* \* \*

---

[1] Research Affiliate, MIT Media Lab.

\* \* \*

## INTRODUCTION

Machine learning is a computational process. To that end, it is inextricably tied to computational power - the tangible material of chips and semiconductors that the algorithms of machine intelligence operate on. Most obviously, computational power and computing architectures shape the speed of training and inference in machine learning, and therefore influence the rate of progress in the technology. But, these relationships are more nuanced than that: hardware shapes the methods used by researchers and engineers in the design and development of machine learning models. Characteristics such as the power consumption of chips also define where and how machine learning can be used in the real world.

In a broader perspective, computational power is also important because of its specific geographies. Semiconductors are designed, fabricated, and deployed through a complex international supply chain. Market structure and competition among companies in this space influence the progress of machine learning. Moreover, since these supply chains are also considered significant from a national security perspective, hardware becomes an arena in which government industrial and trade policy has a direct impact on the fundamental machinery necessary for artificial intelligence (AI).

Despite this, many analyses of the social impact of the current wave of progress in AI have not substantively brought the dimension of hardware into their accounts. While a common trope in both the popular press and scholarly literature is to highlight the massive increase in computational power that has enabled the recent breakthroughs in machine learning, the analysis frequently goes no further than this observation around magnitude.

This paper aims to dig more deeply into the relationship between computational power and the development of machine learning. Specifically, it examines how changes in computing architectures, machine learning methodologies, and supply chains might influence the future of AI. In doing so, it seeks to trace a set of specific relationships between this underlying hardware layer and the broader social impacts and risks around AI. On one hand, this examination shines a spotlight on how hardware works to exacerbate a range of concerns around ubiquitous surveillance, technological unemployment, and geopolitical conflict. On the other, it also highlights the potentially significant role that shaping the development of computing power might play in addressing these concerns.

Part I will examine the role that computational power has played in the progress of machine learning, arguing that its impact has been somewhat flattened in recent accounts looking at the social impact of the technology. Part II will look at trends towards increasing specialization in the hardware used for machine learning, and its implications for control and privacy in the space. Part III will look at the semiconductor supply chain, and its implications for the geopolitics of machine learning. Part IV will examine research developments changing the balance between data and computational power in the workflow of machine learning, and its influence on the economic impact of the technology. It will then conclude with some remarks on the potential role of hardware as a lever for policy action in the space.

**PART I: MACHINE LEARNING AND COMPUTATIONAL POWER**

AI has historically moved through multiple cycles of progress and optimism followed by setbacks and pessimism, so called "AI winters".[2] Present-day excitement around AI, and more specifically the recent breakthroughs in the subfield of machine learning, represent only the latest upswing in this historical pattern.

Machine learning itself, the study of algorithms which improve themselves through data, is not a new domain of research. The fundamentals underlying the modern advances in the field were established by researchers in the 1950s and developed throughout the subsequent decades.[3]

However, neural networks - the specific technique of machine learning driving much of the commercial interest in AI today - were still considered a niche area of research only until relatively recently. As one popular account has put it, "for much of its history most computer scientists saw it [neural networks] as vaguely disreputable, even mystical."[4] It was recognized early in this history that the neural networks proposed during the 1950s and 1960s were limited by the comparatively minimal processing power available at the time.[5] The continued growth of computational power, along with the accumulation of large datasets during the 1990s and 2000s, played a major role in revitalizing progress in neural networks and motivating significant investment within the field of AI more broadly.

The field of computer vision, which focuses on advancing the ability for machines to extract understanding from images and video, offers one representative example on this point. "Traditional" approaches to these tasks in the 1990s and early 2000s focused on algorithms which specified a set of defined features that would be used to process and classify an image.[6]

---

[2] For an account of this history, *see generally* John Markoff, Machines of Loving Grace: The Quest for Common Ground Between Humans and Robots (2015).

[3] *See, e.g.,* B. Farley & W. Clark, *Simulation of self-organizing systems by digital computer*, 4 Transactions of the IRE Professional Group on Information Theory 76–84 (1954).

[4] Gideon Lewis-Kraus, *The Great A.I. Awakening*, The New York Times, December 14, 2016, https://www.nytimes.com/2016/12/14/magazine/the-great-ai-awakening.html (last visited Mar 20, 2018).

[5] *See* Marvin Minsky & Seymour A. Papert, Perceptrons: An Introduction to Computational Geometry (1969) (noting these limitations).

[6] *See, e.g.,* D.G. Lowe, *Object recognition from local scale-invariant features*, 1150–1157 vol.2 (1999), http://ieeexplore.ieee.org/document/790410/ (last visited Mar 20, 2018) (describing the SIFT algorithm, one representative approach).

Neural networks, in contrast, learn the relevant features for classification rather than having them pre-designed into the algorithm.[7]

Expanding computational power and the availability of data changed the practice of computer vision. On the data front, the growth of the consumer web produced a vast library of images for machine learning systems to train on. ImageNet, an annotated dataset of 14 million images in 20 thousand categories assembled by workers on the Amazon Mechanical Turk platform, provided a common dataset for researchers to work with.[8] Computational power also continued to increase in the 2000s, rising from 37 million transistors per chip in 2000 to 2.3 billion transistors per chip by 2009.[9] This was augmented by the finding that a particular kind of computational architecture - the GPU - was particularly well-suited as a platform for neural networks, a development discussed in more detail in Part II.[10]

Leveraging both of these assets, neural networks were able to significantly surpass the performance of earlier techniques in the space.[11] The ImageNet Large Scale Visual Recognition Challenge is a commonly cited marker of this transition. Hosted since 2010, the Challenge brings together researchers to compete in designing systems to solve a set of visual recognition tasks.[12] Performance in 2010 and 2011, which featured teams using traditional techniques in the space, were never able to reduce error rates below 25%, with most teams showing much higher rates of error.[13]

---

[7] *See, e.g.,* Alex Krizhevsky, Ilya Sutskever & Geoffrey E. Hinton, *ImageNet classification with deep convolutional neural networks*, 60 Communications of the ACM 84–90 (2017) (demonstrating this learned feature approach).
[8] Dave Gershgorn, The data that transformed AI research—and possibly the world Quartz, https://qz.com/1034972/the-data-that-changed-the-direction-of-ai-research-and-possibly-the-world/ (last visited Mar 20, 2018).
[9] Moore's Law: Transistors per microprocessor, Our World in Data, https://ourworldindata.org/grapher/transistors-per-microprocessor (last visited Mar 20, 2018).
[10] *See* Rajat Raina, Anand Madhavan & Andrew Y. Ng, *Large-scale deep unsupervised learning using graphics processors*, 1–8 (2009), http://portal.acm.org/citation.cfm?doid=1553374.1553486 (last visited Mar 20, 2018).
[11] Cf. From not working to neural networking, The Economist, 2016, https://www.economist.com/news/special-report/21700756-artificial-intelligence-boom-based-old-idea-modern-twist-not (last visited Mar 20, 2018).
[12] ImageNet Large Scale Visual Recognition Challenge (ILSVRC), http://www.image-net.org/challenges/LSVRC/ (last visited Mar 20, 2018).
[13] Id.

AlexNet, a system submitted to the competition by researchers Geoffrey Hinton, Ilya Sutskever, and Alex Krizhevsky in 2012, was both the first entry to apply neural networks in the Challenge and the first to achieve a below 25% error rate.[14] The excitement around these results, and the margin of improvement over established techniques, led one researcher to state that the "Imagenet 2012 event was definitely what triggered the big explosion of AI today".[15]

Computational power has for this reason been fundamental to the present-day breakthroughs in machine learning. Even if the necessary data been widely available at an earlier point historically, a lack of computational power would have effectively prevented neural networks from achieving their current level of performance.

*The Tropes of Computational Power*

The narrative of computational power and machine learning typically ends here. Recent pieces examining these technological breakthroughs have often focused more on the implications of what the technology can do, rather than the implications of how it is being done.[16] To the extent that computational power is mentioned, it is typically addressed simply as an enabling factor in the emergence of machine learning. The prevailing attribute highlighted in these accounts has tended to be one of magnitude: the processing power of the chips running machine learning have been seen as their primary contribution.

The 2016 White House report *Preparing for the Future of Artificial Intelligence* is illustrative. The paper focuses on computational power only in passing as one of the three factors enabling the present-day breakthroughs in machine learning. "[T]he availability of *big data*…dramatically *improved machine learning approaches and algorithms*…the capabilities of *more powerful computers*."[17] An accompanying paper released at the same time, *The National Artificial Intelligence Research and Development Plan,* highlights improved hardware for machine learning as a priority, but only to the extent that chips with higher levels of performance are needed to drive the technology forwards.[18] Other reports from the European Union, civil society groups, and researchers on the topic of AI have followed a similar set of themes when considering the role of computational power.[19]

---

[14] Id.
[15] *See supra* note 8 (for a visualization of these results).
[16] *See, e.g.,* JURI Committee, European Civil Law Rules in Robotics (2016), *available at* http://www.europarl.europa.eu/RegData/etudes/STUD/2016/571379/IPOL_STU(2016)571379_EN.pdf; Urs Gasser, AI and the Law: Setting the Stage, Medium (2017), https://medium.com/berkman-klein-center/ai-and-the-law-setting-the-stage-48516fda1b11 (last visited Mar 20, 2018).
[17] White House National Science and Technology Council, Preparing for the Future of Artificial Intelligence 6 (2016), *available at*

This narrative perhaps leaves out an important part of the story. Such a shorthand enables a focus on the numerous problematic ways that machine learning might be applied and the implications of those applications for justice, equity, and a host of other values. However, it also flattens out the role that computational power plays in these issues to simply that of a trigger for technological progress. This may miss the significant and nuanced ways that hardware influences the impact of AI on these broader values and social concerns.

Computational power does more than simply make the present-day breakthroughs in machine learning possible. The medium is a significant message here: hardware actively shapes the landscape of what can be done with the technology of machine learning, and plays a significant role in influencing how it will evolve going forwards.

The contours of computational power play a role in defining who has control over and access to the benefits of machine learning, and the actors that will play a role in its governance. It plays a role in the politics of the technology, both at the level of an individual citizen and in the broader competition between states. Computational power, in defining the speed at which machine learning models may trained and experimented on, shapes the speed at which the technology advances and therefore serves to define

---

https://obamawhitehouse.archives.gov/sites/default/files/whitehouse_files/microsites/ostp/NSTC/preparing_for_the_future_of_ai.pdf.

[18] White House National Science and Technology Council, The National Artificial Intelligence Research and Development Strategic Plan 21 (2016).

[19] *See e.g.,* Royal Society (Great Britain), Machine learning: the power and promise of computers that learn by example (2017) *available at* https://royalsociety.org/~/media/policy/projects/machine-learning/publications/machine-learning-report.pdf; House of Commons Science and Technology Committee, Robotics and Artificial Intelligence (2016), https://publications.parliament.uk/pa/cm201617/cmselect/cmsctech/145/145.pdf; World Economic Forum, Assessing the Risk of Artificial Intelligence, Global Risks Report 2017, http://wef.ch/2izSQRP (last visited Mar 22, 2018); David Bollier, Artificial Intelligence: The Great Disruptor, Aspen Institute (2018) *available at* http://csreports.aspeninstitute.org/documents/AI2017.pdf.

its broader economic impact. These impacts turn on more than simply the amount of processing power available, but on the details of computational architecture, supply chains, and the co-evolution of the machine learning field itself.

To that end, the evolving research and commercial ecosystem around hardware is more than just a sideshow: shifts in these underlying technologies have a significant place in understanding the impact of AI on society as a whole. Parts II, III, and IV work to map these many connections by drawing the lines from the changing landscape of computational power to the bigger social challenges surrounding AI.

**PART II: SPECIALIZATION IN COMPUTATIONAL POWER**

Computational power is not a simple matter of magnitude. The specific architecture of a chip plays a major role in determining whether or not it is effective in dealing with a given computational task. In general, the industry has tended towards increasingly specialized platforms for machine learning as the field continues to grow and attract commercial interest. In this sense, hardware has moved in a direction opposite to software: a shift towards narrower specialization in chips has proceeded even as the research field has been focused on building ever more general learning systems.

Two dynamics shape this marketplace for machine learning hardware. One is an inverse relationship between performance and flexibility.[20] While general purpose computing power can take on a wide range of tasks and can be easily configured to take on new tasks, it tends to be outpaced by hardware which is built for a specific purpose.[21] However, this increased performance comes at a cost: specialized hardware accommodates a relatively smaller set of use cases and has an architecture which is less easy to change after it is deployed.[22] These specialized platforms are also frequently more expensive than commodified general platforms.[23] One overarching question is whether the demand for machine learning driven products and the research community will tend over time to favor architectures that are more or less flexible given this tradeoff.

---

[20] *See generally* Inside the Microsoft FPGA-based configurable cloud, Channel 9, https://channel9.msdn.com/Events/Build/2017/B8063 (last visited Mar 20, 2018) (discussing these trade-offs).
[21] *See* Griffin Lacey, Graham W. Taylor & Shawki Areibi, *Deep Learning on FPGAs: Past, Present, and Future*, arXiv:1602.04283, 6 (2016), http://arxiv.org/abs/1602.04283 (last visited Feb 13, 2018) (discussing cost differentials).
[22] Id.
[23] Id.

A second important dynamic is that the hardware for *training* a machine learning model to accomplish a task can differ significantly from the hardware used to conduct *inference* with an already trained model. This is due to the different demands at each step of the machine learning workflow. For instance, energy consumption may matter for a computer vision system operating on a mobile device, though it may not matter when that computer vision system is being trained initially in a data center.[24] Latency - the time delay between input and output of a system - might be a significant factor in a high-speed navigation context, where speed of inference would reduce the time needed for a course correction.[25] However, as with energy consumption, latency may not prove to be as significant when the navigation system is being trained. These considerations influence what kinds of hardware are used at which points in the lifecycle of a machine learning system. They can be viewed as separate though overlapping markets, with hardware platforms being offered either for training or inference, and some offering support for both.[26]

Background: CPUs to GPUs

Graphics processing units (GPUs) form the present-day backbone of the machine learning workflow.[27] GPUs are the primary platform for both training and inference, and are widely used both for basic research and in the practical development and deployment of machine learning driven products in the marketplace.[28]

The outsize role that GPUs play in machine learning is the result of an unexpected historical convergence. As their name suggests, GPUs were originally designed to support computer graphics and image processing applications.[29] To accomplish this, GPUs feature an architecture which distributes computational tasks across a large number of cores to be processed in parallel.[30] This is in contrast to central processing units (CPUs), which feature a smaller number of more powerful cores that are optimized for handling just a few tasks simultaneously.[31]

---

[24] *See* Vivienne Sze et al., *Efficient Processing of Deep Neural Networks: A Tutorial and Survey*, arXiv:1703.09039, 5-6 (2017), http://arxiv.org/abs/1703.09039 (last visited Feb 13, 2018).
[25] Id at 26.
[26] *See, e.g.,* Jeff Dean and Urs Hölzle, Build and train machine learning models on our new Google Cloud TPUs, Google (2017), https://www.blog.google/topics/google-cloud/google-cloud-offer-tpus-machine-learning/ (last visited Mar 20, 2018) (earlier generations of Google's specialized machine learning chips were focused on inference, with the latest version supporting both inference and training).
[27] *See* Deloitte, Hitting the Accelerator: The Next Generation of Machine-Leaning Chips (2017), *available at* https://www2.deloitte.com/content/dam/Deloitte/global/Images/infographics/technologymediatelecommunications/gx-deloitte-tmt-2018-nextgen-machine-learning-report.pdf.
[28] Id. at 1.

This parallel architecture allows the GPU to be uniquely well suited for machine learning applications. At its root, neural network training and inference relies on the execution of a large number of identical matrix multiplication calculations.[32] This uniformity enables these operations to be parallelized and distributed across the many cores offered by the GPU.[33] This enables the GPU to outperform CPU architectures which have comparatively more powerful processors but manage tasks in a more serial format.[34]

Increasing Specialization: FPGAs and ASICs

The repurposing of GPUs as the primary hardware platform for machine learning reflects a selection among available technologies. As interest in machine learning has continued to grow, so has the notion of developing hardware entirely purpose-built for these applications become more attractive. Discussion within the industry has focused on the possibility of using field-programmable gate arrays (FPGAs) and application-specific integrated circuits (ASICs) as the next primary platforms for machine learning.[35]

FPGAs are distinct from CPUs and GPUs in that they do not run programs in stored memory. Instead, they are collection of standardized "logic blocks" whose relationships can be configured by a programmer once the chip is received from a manufacturer.[36] ASICs are purpose-built chip boards which are specific to a purpose and cannot be easily reconfigured after they are manufactured.[37]

---

[29] Id. at 2.
[30] *See supra* note 10.
[31] Id.
[32] *See supra* note 22 at 12.
[33] *See supra* note 10.
[34] *See* Cade Metz, The Race To Build An AI Chip For Everything Just Got Real, WIRED, https://www.wired.com/2017/04/race-make-ai-chips-everything-heating-fast/ (last visited Feb 13, 2018) (describing some of the limitations of the CPU).
[35] *See, e.g.,* Karl Freund, Will ASIC Chips Become The Next Big Thing In AI?, Forbes, https://www.forbes.com/sites/moorinsights/2017/08/04/will-asic-chips-become-the-next-big-thing-in-ai/ (last visited Mar 20, 2018).

FPGAs and ASICs are particularly attractive in the context of machine learning inference. Both devices consume less energy than CPUs and GPUs, and their specialization allows for greater speed.[38] These gains come at the loss of flexibility and an increased cost. FPGAs and ASICs cannot be as easily and quickly configured to run a wide range of tasks. Both are relatively more expensive when compared with CPUs and GPUs. This is particularly the case with ASICs, which are "bespoke" projects that are expensive and time-consuming to produce. This makes them cost-effective as a platform only in significant quantities.[39]

However, these benefits may outweigh the costs, particularly in circumstances where a specific kind of machine learning inference is reliably needed at a mass scale. Project Catapult, an initiative launched by Microsoft, has shown high performance for FPGAs as the core computing unit in their data centers.[40] FPGAs and ASICs have also been considered a promising approach in the autonomous vehicles context, where the tasks a machine learning system will need to take on will be relatively stable and where chips will be needed for a large number of vehicles.[41]

---

[36] *See* What is an FPGA? Field Programmable Gate Array, https://www.xilinx.com/products/silicon-devices/fpga/what-is-an-fpga.html (last visited Mar 20, 2018).

[37] Jeff Dean, Machine Learning for Systems and Systems for Machine Learning, NIPS 2017, *available at* http://learningsys.org/nips17/assets/slides/dean-nips17.pdf (noting the design challenges with ASICs).

[38] *See supra* notes 19-20.

[39] Id.

[40] *See* Kalin Ovtcharov et al., *Accelerating deep convolutional neural networks using specialized hardware*, Microsoft Research Whitepaper, 2 (2015), *available at* https://www.microsoft.com/en-us/research/wp-content/uploads/2016/02/CNN20Whitepaper.pdf.

[41] *See, e.g.,* ASIC might be the mainstream chip for autonomous driving, a chance for Chinese start-ups, VehicleTrend, https://www.vehicle-trend.com/Knowledge/20180108-1131.html (last visited Mar 21, 2018); Phil Kalaf, Self-Driving Cars, Wireless Data? It's Time to Thank the Humble FPGA. IDS (2017), http://www.idsforward.com/wireless-data-thank-humble-fpga/ (last visited Mar 21, 2018); Yu Wang et al, Reconfigurable Processor for Deep Learning in Autonomous Vehicles, https://www.itu.int/en/journal/001/Documents/itu2017-2.pdf (2017); Harsh Chauhan, Can Intel Dominate This Market by Overcoming This Smaller Rival? The Motley Fool (2017), https://www.fool.com/investing/2017/11/24/can-intel-dominate-this-market-by-

While FPGAs and ASICs seem to show promise in the inference context, they have traditionally had some limitations that have made them less attractive as platforms for training. For one, FPGAs and ASICs have tended to be less accurate, relying on "fixed point" computation or featuring comparatively limited floating-point performance.[42] This has made them comparatively limited in performing the accurate level of calculation needed in the training process. These devices have also had limited external memory bandwidth, preventing them from efficiently conducting the matrix multiplication needed for training.[43]

However, the potential speed and energy consumption gains presented by FPGA and ASIC architectures have encouraged research which appears to be eliminating some of these limitations over time. In 2017, Intel researchers released software which maximizes data reuse and minimizes external memory bandwidth to boost training performance on FPGAs.[44] The latest generation of Google's "Tensor Processing Unit" (TPU), a specialized ASIC, supports both training and inference.[45] The claimed improvements are quite significant. One recent talk from Google in 2017 noted that the TPU ASIC was able to execute training tasks at ten to fourteen times the speed of their previous production setups with a relatively smaller number of machines.[46]

Moving Forwards

It remains unclear whether or not more specialized, less flexible hardware will unseat the preeminent place of the GPU in machine learning training and inference. Producers of FPGAs and ASICs are releasing performance benchmarks showing significant improvements over the GPU for both kinds of tasks. Google claims that its TPU ASIC is able to conduct

---

overcoming-this.aspx (last visited Feb 20, 2018) (noting the application of FPGAs in the autonomous vehicles context).

[42] *See* Brian Bailey, Machine Learning's Growing Divide, Semiconductor Engineering, https://semiengineering.com/machine-learnings-growing-divide/ (last visited Mar 12, 2018) (reviewing these issues); Understanding Peak Floating-Point Performance Claims, https://www.altera.com/en_US/pdfs/literature/wp/wp-01222-understanding-peak-floating-point-performance-claims.pdf (last visited Feb 15, 2018) (reviewing the floating point issues in more depth).

[43] Id.

[44] Utku Aydonat et al., *An OpenCL(TM) Deep Learning Accelerator on Arria 10*, arXiv:1701.03534 [cs] (2017), http://arxiv.org/abs/1701.03534 (last visited Mar 12, 2018).

[45] *See supra* note 34.

[46] *See supra* note 34 (describing these improvements).

inference fifteen to thirty times faster than contemporary GPUs and CPUs.[47] In a similar vein, Graphcore - one prominent startup focusing on specialized machine learning hardware - claims that eight of its proprietary "IPU" cards are equivalent to 128 contemporary GPUs.[48]

Despite this, benchmarking issues persist and it is challenging to evaluate these claims in a systematic way.[49] For its part, GPU leader Nvidia has challenged the performance claimed by Google of its TPUs, noting that it failed to compare its chips against its latest generation of hardware.[50] At the moment, the semiconductor industry does not yet have a common scheme for evaluating the performance of machine learning specialized hardware as it does in the CPU space.[51] This is significant because the specific architecture of the neural network and how it is implemented can have a significant impact on reported performance.[52]

Though some commentators have framed the industry choices between GPUs, FPGAs, and ASICs as a mutually exclusive ones, it is not clear that this will be the case in practice.[53] Even if FPGA and ASIC designs do not ultimately become a new standard for training and inference in machine learning writ large, it seems likely that they will become a natural option for certain applications of machine learning systems, particularly in the consumer products context. Recent moves by industry leaders seems to recognize this reality. Despite its leadership and championing of a GPU-focused model, Nvidia's latest Drive PX product features a specialized "Deep Learning Accelerator" (DLA) module as it attempts to cater to applications in autonomous vehicles.[54] Nvidia has also open-sourced its designs for the DLA, a move likely to drive down the cost of this specialized hardware going forwards by enabling others to manufacture the same designs.[55]

---

[47] Norman P. Jouppi et al, In-Datacenter Performance Analysis of a Tensor Processing Unit, https://arxiv.org/ftp/arxiv/papers/1704/1704.04760.pdf (last visited Feb 15, 2018).
[48] Graphcore Benchmarks, Presentation at NIPS 2017, *available at* https://cdn2.hubspot.net/hubfs/729091/NIPS2017/NIPS%2017%20-%20benchmarks%20final.pdf?t=1521107772551.
[49] *See supra* note 22, at 26-27 (describing the many influences on chip performance).
[50] Jensen Huang, AI Drives the Rise of Accelerated Computing in Data Centers, Nvidia Blog, https://blogs.nvidia.com/blog/2017/04/10/ai-drives-rise-accelerated-computing-datacenter/ (Apr 10, 2017).
[51] *See, e.g.,* SPEC CPU 2017, https://www.spec.org/cpu2017/.
[52] *See supra* note 22, at 26-27.
[53] *See, e.g.,* FPGA Based Deep Learning Accelerators Take on ASICs, The Next Platform (2016), https://www.nextplatform.com/2016/08/23/fpga-based-deep-learning-accelerators-take-asics/ (last visited Feb 13, 2018); Does the future lie with CPU+GPU or CPU+FPGA?, Scientific Computing World, https://www.scientific-computing.com/news/analysis-opinion/does-future-lie-cpugpu-or-cpufpga (last visited Feb 20, 2018).

These trends provide a framework for thinking about the economics of various machine learning applications, and how and where the technology might be used in practice. To that end, it begs the question of how these shifts in hardware specialization might influence the overall impact of machine learning on society and the governance of the technology.

*Impact: The Geography of Training and Inference*

Machine learning is not an abstract force, but a computational task that takes place *somewhere*. Hardware capabilities and the particular economics of processors are critical since they define the spatial dimensions of machine learning and what it likely to be applied towards.

Power consumption defines whether or not machine learning computation can be done on a small, mobile device, or must have access to a reliable and continuous power source. High energy costs limit the ability to embed machine learning systems directly on a device. In the very least, it limits the application of machine learning to situations with sufficient connectivity for a device to communicate with a larger pool of computational power hosted in the "cloud."[56]

Latency is also crucial in this respect. Even if a chip is able to operate at low power on a mobile device, it may be ineffectual for a particular use in the field if it is insufficiently responsive for the intended purpose. This acts as a bar to certain real-time or mission-critical uses of machine learning where an alternative cloud architecture would also produce similarly unacceptable levels of latency.

---

[54] Karl Freund, Why Nvidia is Building Its Own TPU, Forbes, https://www.forbes.com/sites/moorinsights/2017/05/15/why-nvidia-is-building-its-own-tpu (last visited Mar 20, 2017).
[55] *See* Tom Simonite, To Compete With New Rivals, Chipmaker Nvidia Shares Its Secrets, WIRED, Sept 29, 2017, https://www.wired.com/story/to-compete-with-new-rivals-chipmaker-nvidia-shares-its-secrets/ (last visited Feb 20, 2018).
[56] *See supra* note 22, at 5-6 (discussing these different configurations).

Power consumption and latency are barriers to the application of machine learning within certain domains. This includes the placement of these systems on small devices with an untethered power source, and in low bandwidth situations with poor connectivity. The rise of ASICs and FPGAs - as well as ongoing improvements to GPUs - suggest that machine learning hardware will erode these limitations on the placement of machine learning systems over time, particularly for mass produced consumer products that have economies of scale.

The Geography of Inference

These developments offer a mixed blessing to those concerned about the harmful possibilities of machine learning inference. Machine learning can increasingly be integrated into a range of different products and services and used in situations where it was previously considered impractical to do so. For civil libertarians, FPGAs and ASICs enable the expansion of machine learning as a means of conducting surveillance: small, low power devices can now incorporate the advances of computer vision to recognize people and objects even in areas with low bandwidth. For those worried about the misuse of machine learning by "bad actors", specialized hardware makes it more possible to benefit from the technology without reliance on cloud services where harmful activity might be more easily tracked and halted.[57]

Also concerning is the fact that the more inflexible architectures of FPGAs and ASICs might potentially make it more challenging to repair machine learning systems when flaws are discovered. A growing body of research continues to highlight the point both that machine learning systems frequently can render biased, discriminatory results, and are potentially vulnerable to malicious manipulation.[58] Where a trained machine learning model is "hard wired" into a chip, the discovery that it has these flaws may make repair a more expensive and protracted process as it requires a replacement of the processor itself, rather than modification of software. This problem applies with particular force in an "embedded" setting where chips are sold and distributed with a product and there is no centralized means of changing their behavior once they have left the factory.

---

[57] For a review of these concerns, *see, e.g.,* Miles Brundage et al, The Malicious Use of Artificial Intelligence (2018), available at https://maliciousaireport.com/.

[58] *See, e.g.,* Joy Buolamwini & Timnit Gebru, *Gender Shades: Intersectional Accuracy Disparities in Commercial Gender Classification*, PMLR 81:1–15, 2018, *available at* http://proceedings.mlr.press/v81/buolamwini18a/buolamwini18a.pdf; Solon Barocas & Andrew D. Selbst, Big Data's Disparate Impact (2016), https://papers.ssrn.com/abstract=2477899 (last visited Mar 21, 2018); Ian Goodfellow et al, Attacking Machine Learning with Adversarial Examples, OpenAI Blog (2017), https://blog.openai.com/adversarial-example-research/ (last visited Mar 21, 2018); Tom B. Brown et al., *Adversarial Patch*, arXiv:1712.09665 [cs] (2017), http://arxiv.org/abs/1712.09665 (last visited Mar 21, 2018).

At the same time, FPGAs and ASICs also raise the possibility that machine learning may be architected in a more robustly privacy-protecting ways going forwards. Since specialized computing power enables machine learning inference to be done on the device itself, it also opens the possibility that machine learning capabilities might be provided without ever having personal data leave a device. Consider a computer vision system which helps users quickly sort through their photos to find friends and family members. At present, the energy costs of inference might require that these photos be uploaded from a smartphone to a central server to be processed and tagged.[59] Specialized hardware might enable an alternative architecture in which the machine learning model is embedded on the smartphone itself, such that the photos themselves do not need to be shared with a third-party to be analyzed by the system.[60]

This remains up the air as FPGAs and ASICs for machine learning enter the scene and attempt to find viable niches in the application of the technology. The economics of these hardware platforms will influence the viability of alternative architectures in various markets, and in doing so will inform whether advocates and policymakers are able to argue for the feasibility of more privacy sensitive approaches to machine learning going forwards.

The Geography of Training

It is important to recognize that the geography of training may look quite different from the geography of inference.[61] As discussed above, FPGAs and ASICs have been traditionally somewhat limited as platforms for the training of machine learning systems. While the possibilities of using specialized hardware for training continue to be developed by Google and other companies, the reality in the near-term seems to be that training will remain the province of the GPU for many researchers and practitioners. Combined with the fact that training is likely to continue being computationally intensive for the foreseeable future, it is likely that the creation of machine learning models will continue to happen within centralized data centers.

---

[59] *See supra* note 22 at 5-6.
[60] *See, e.g.,* Ben Popper, Google's new Clips camera is invasive, creepy, and perfect for a parent like me The Verge (2017), https://www.theverge.com/2017/10/5/16428708/google-clips-camera-privacy-parents-children (last visited Mar 21, 2018) (describing an architecture along these lines).
[61] Nvidia, GPU-Based Deep Learning Inference: A Performance and Power Analysis, 4-5 (2015), available at https://www.nvidia.com/content/tegra/embedded-systems/pdf/jetson_tx1_whitepaper.pdf (noting the differing demands of training and inference).

FPGAs and ASICs therefore seem likely to have a differential impact on the geography of machine learning. In the past, pre-existing data center infrastructure, business models, and the energy consumption of existing processors tended to encourage an architecture where training and inference were situated in the same, centralized locations. As this specialized hardware matures, it seems likely to encourage a more distributed pattern in inference, permitting the application of machine learning "on device" in a broader set of contexts. At the same time, the continuing computational costs and energy requirements of training mean that the creation of machine learning systems will continue to largely take place in a relatively smaller number of central facilities.

This geographic pattern has implications for the governance of machine learning. Training of the most complex, sophisticated models will continue to take place in a small number of locations among the set of actors who have the financial resources to maintain or rent the computational power necessary. However, once trained, machine learning models can be increasingly diffused and distributed. No doubt some types of machine learning models will continue to be offered "as a service," with inference taking place in the cloud. However, FPGAs and ASICs open the door to inference no longer being tethered in this particular way. Simultaneously, these platforms - ASICs in particular - are more inflexible, making modifications after distribution more difficult.

Whereas in the past it was more possible to repair flaws in machine learning models after training and deployment by directly modifying the model provided to many endpoints through the cloud, the specialization of hardware suggests an environment where these harms may be more challenging to rectify in the post-training phase. "Hard-wired" chips may be difficult to recall quickly, or otherwise difficult to modify when in the field. This may put increased pressure on companies to engage in more thorough pre-deployment checks and verification on these systems, rather than adopting a development approach that takes a more "launch-and-iterate" stance. From a governance perspective, these less easily rectifiable downstream harms may push regulators towards an approach that puts a growing set of responsibilities on the entities creating and providing platforms for creating machine learning models to take precautions prior to wider distribution.

It is worth recognizing that the continued development of the machine learning field may alter this balance over time. Progress continues to be made in the subfield of federated learning - which envisions an architecture in which many independent, distributed processors train locally and share updates to a model with one another.[62] This work may become more practically feasible to implement as ASICs for machine learning mature and increasingly allow embedded training to happen on a device. Breakthroughs in one-shot learning, which would enable the effective training of models with a relatively smaller number of examples, might also lower the computational bar to executing training tasks in a more distributed way.[63]

## PART III: SUPPLY CHAIN AND COMPUTATIONAL POWER

CPUs, GPUs, FPGAs, and ASICs are all ultimately products in a complex global supply chain for semiconductors. Beyond simply connecting changing computational architectures to the social impact of machine learning as we did in the previous section, we can dig deeper to examine how the commercial specifics of its manufacture also has broader implications.

Both the geography of semiconductor manufacturing and its place as a strategic asset in the context of national security make it likely that computing power will become an important arena in the geopolitics of AI. This seems to be particularly the case as China increasingly invests in becoming a leader in machine learning while continuing an ongoing effort to reshape the global semiconductor industry.

<u>The Semiconductor Supply Chain</u>

The hardware platforms discussed in Part II are just one facet of the much broader industry for semiconductors. Semiconductor chips, "tiny electronic device[s] comprised of billions of components that store, move, and process data" are the "enabling technology of the information age."[64] These chips give computers the power to run software applications, and are the key building block for a range of other devices "from cell phones and gaming systems to aircraft and industrial machinery to military equipment and weapons."[65] Not surprisingly given their broad scope of application, semiconductors are a massive global industry. In 2015, worldwide semiconductor sales were $335 billion, growing 15% since 2012.[66]

---

[62] *See, e.g.,* Jakub Konečný et al., *Federated Optimization: Distributed Machine Learning for On-Device Intelligence*, arXiv:1610.02527 [cs] (2016), http://arxiv.org/abs/1610.02527 (last visited Mar 21, 2018).
[63] *See, e.g.,* Adam Santoro et al., *One-shot Learning with Memory-Augmented Neural Networks*, arXiv:1605.06065 [cs] (2016), http://arxiv.org/abs/1605.06065 (last visited Mar 21, 2018).

Many production steps are required to deliver a finished semiconductor chip. Some companies are "integrated device manufacturers", or IDMs, which manage all aspects of semiconductor production from start to finish. This includes design, manufacturing, assembly, testing, and packaging.[67] Companies adopting this model include Intel, Samsung, and Texas Instruments.[68]

However, many businesses specialize only in a particular part of this supply chain, contracting out tasks to other companies in the ecosystem as needed. Of particular importance in the discussion of machine learning hardware is the role of so-called "fabless foundries." These businesses focus on the design of semiconductor chips, and contract out the manufacturing, often called "fabrication," of the final product.[69] Companies adopting this model include AMD, Broadcom, and Qualcomm.[70] Many of the companies leading the development of machine learning specific hardware are "fabless". This allows these businesses to avoid the massive capital outlay and expense of building and maintaining a chip "fab". Building a single advanced plant for fabricating semiconductors can cost up to $20 billion.[71]

---

[64] Michaela D. Platzer and John F. Sargent Jr., U.S. Semiconductor Manufacturing: Industry Trends, Global Competition, Federal Policy, Congressional Research Service 1 (2016) https://fas.org/sgp/crs/misc/R44544.pdf (last visited Feb 13, 2018).
[65] Id.
[66] Id. at 3.
[67] *Cf.* Semiconductor Industry Association, Beyond Borders: The Global Semiconductor Value Chain 7 (2016), available at
https://www.semiconductors.org/document_library_and_resources/trade/beyond_borders_the_global_semiconductor_value_chain/.
[68] Id.
[69] Id.
[70] Id.
[71] TSMC Ready to Spend $20 Billion on its Most Advanced Chip Plant, Bloomberg.com, October 6, 2017, https://www.bloomberg.com/news/articles/2017-10-06/tsmc-ready-to-spend-20-billion-on-its-most-advanced-chip-plant (last visited Mar 20, 2018).

The market for GPUs has been dominated by Nvidia, a fabless foundry. One industry analysis concluded that, as of the third quarter of 2017, Nvidia represented 72.8% of the market share for GPUs, with the rest being controlled by AMD, another fabless foundry.[72] Both are headquartered in Santa Clara, California.

The market for FPGAs has also been dominated by a small set of fabless foundries. In 2016, Xilinx led this segment with a market share of 53%.[73] Altera, another FPGA specialist, was purchased by Intel in 2015 and accounted for 36% of the market.[74] These were distantly followed by competitors Microsemi (7%) and Lattice Semiconductor (3%).[75] This roughly holds stable from the market share in 2015.[76] All of these companies are based in the United States. All but one, Lattice Semiconductor, are headquartered in California.[77]

Estimating market share in the context of specialized machine learning ASICs is more challenging. For one, the market is still emerging: no major player in the space is currently engaging in mass production and public sale of ASICs as a platform for machine learning. At the time of writing, Google is only distributing their TPU ASIC to a relatively small circle of researchers, and offers limited access to TPU computing cycles via its cloud services.[78] In any case, because these chips are highly customized for particular purposes, it may be challenging to eventually define a single "market" which usefully groups together the different types of devices that might be fabricated as an ASIC for machine learning.

---

[72] Harsh Chauhan, Nvidia Is Running Away With the GPU Market The Motley Fool (2017), https://www.fool.com/investing/2017/12/06/nvidia-is-running-away-with-the-gpu-market.aspx (last visited Feb 20, 2018).
[73] And the Winner of Best FPGA of 2016 is... | EE Times, EETimes, https://www.eetimes.com/author.asp?section_id=36&doc_id=1331443 (last visited Feb 20, 2018).
[74] Id.
[75] Id.
[76] Id.
[77] Xilinx, Corporate Locations, https://www.xilinx.com/about/contact/corporate-locations.html; Intel Programmable Solutions Group (PSG) Locations, https://www.altera.com/about/contact/contact/altera-hq.html; Microsemi, Locations, https://www.microsemi.com/locations; Lattice Semiconductor, Locations, http://www.latticesemi.com/About.
[78] Google is giving a cluster of 1,000 Cloud TPUs to researchers for free, TechCrunch (2017), http://social.techcrunch.com/2017/05/17/the-tensorflow-research-cloud-program-gives-the-latest-cloud-tpus-to-scientists/ (last visited Mar 21, 2018); Cloud TPUs - ML accelerators for TensorFlow, Google Cloud, https://cloud.google.com/tpu/ (last visited Mar 21, 2018).

Since the major leaders in machine learning hardware are "fabless", they depend on a separate ecosystem of companies to provide the actual fabrication of the chips they design. These companies, called "pure play foundries" or simply "foundries", are a highly consolidated marketplace. In 2016, Taiwan Semiconductor (TSMC) accounted for 59% of the global market for fabrication.[79] Running significantly behind were GlobalFoundries (11%), United Microelectronics Corporation (UMC) (9%), and Semiconductor Manufacturing International Corporation (SMIC) (6%).[80] TSMC and UMC are based in Taiwan, with SMIC in China and GlobalFoundries in the US.[81] While based in the US, GlobalFoundries is owned by the Emirate of Abu Dhabi through its state-owned investment arm Advanced Technology Investment Company (ATIC).[82]

There exists a network of somewhat stable relationships between this handful of leading "fabless" foundries that are designing much of the hardware that machine learning takes place on, and the small number of companies that do their manufacturing. In the GPU space, Nvidia contracts much of its high-performance GPU production to TSMC.[83] In 2009, GlobalFoundries was spun-off from AMD as part of a transition of the latter towards a "fabless" model.[84] As a result of this historical relationship, AMD has traditionally worked closely with GlobalFoundries, though recently

---

[79] Pure-Play Foundry Market Surges 11% in 2016 to Reach $50 Billion!, , http://www.icinsights.com/news/bulletins/PurePlay-Foundry-Market-Surges-11-In-2016-To-Reach-50-Billion/ (last visited Feb 20, 2018).
[80] Id.
[81] TSMC, Business Contacts, http://www.tsmc.com/english/aboutTSMC/business_contacts.htm; UMC, Locations, http://www.umc.com/english/contact/index.asp; SMIC, About Us, http://www.smics.com/eng/about/about.php; GlobalFoundries, About Us, https://www.globalfoundries.com/about-us.
[82] Mark LaPedus, ATIC takes control of GlobalFoundries | EE Times, EETimes, https://www.eetimes.com/document.asp?doc_id=1258215 (last visited Mar 21, 2018).
[83] *See* Nvidia: TSMC will remain a 'very important' foundry partner, KitGuru (2015), https://www.kitguru.net/components/graphic-cards/anton-shilov/nvidia-tsmc-will-remain-a-very-important-foundry-partner/ (last visited Feb 20, 2018); Ashraf Eassa, NVIDIA Corp.'s Relationship With Taiwan Semiconductor Manufacturing Is Deepening The Motley Fool (2017), https://www.fool.com/investing/2017/05/17/nvidia-corp-relationship-taiwan-semiconductor.aspx (last visited Mar 21, 2018).
[84] Benjamin Pimentel, GlobalFoundries created from AMD spin-off MarketWatch, https://www.marketwatch.com/story/globalfoundries-created-amd-spin-off-the (last visited Mar 21, 2018).

announced that they will be splitting their new GPU production between them and TSMC.[85]

FPGA production operates with a slightly different set of connections between chip designers and associated foundries. Xilinx has worked in the past with UMC though increasingly partners with TSMC on its more recent hardware.[86] Altera was acquired by Intel in 2015 and relies in part on its corporate parent for fabrication services.[87] Microsemi announced in 2013 that it too would work with Intel for its fabrication needs.[88] Lattice Semiconductor works with both UMC and TSMC, along with some smaller foundries.[89]

The geographic distribution of these players is mirrored in the overall structure of the industry. US firms account for the largest share of the global market, accounting for 50% of semiconductor sales in 2016.[90] However, the actual fabrication of semiconductor devices largely takes place outside of the United States. In 2015, about three-quarters of the world's advanced semiconductor fabrication capacity was located in South Korea, Taiwan, Japan, and China.[91] This continues a historical trend of production capacity moving from the US to the Asia-Pacific region. In 1980 the US accounted for 42% of global manufacturing capacity, a number which dropped consistently over subsequent years to 16% by 2007.[92]

<u>US National Security and the Semiconductor Industry</u>

---

[85] *See* AMD are splitting 7nm Zen 2 CPU and Vega GPU manufacturing between TSMC and GloFo | PCGamesN, , https://www.pcgamesn.com/amd-7nm-tsmc-globalfoundries (last visited Feb 20, 2018).
[86] Dylan McGrath, Xilinx confirms: Samsung, TSMC in, UMC out at 28-nm | EE Times, EETimes, https://www.eetimes.com/document.asp?doc_id=1173112 (last visited Mar 21, 2018).
[87] Can Intel Dethrone The Foundry Giants?, Semiconductor Engineering, https://semiengineering.com/intel-dethrone-foundry-giants/ (last visited Mar 21, 2018).
[88] Microsemi, Microsemi Selects Intel Corporation Foundry Services for the Development of Digital Integrated Circuits, https://investor.microsemi.com/2013-05-01-Microsemi-Selects-Intel-Corporation-Foundry-Services-for-the-Development-of-Digital-Integrated-Circuits (last visited Mar 20, 2018).
[89] Lattice Semiconductor, ISO 9001/ TS 16949/ ISO 14001 Certificates, http://www.latticesemi.com/en/Support/QualityAndReliability/ManufacturingPartners/ISOCertificates (last visited Mar 20, 2018).
[90] *See supra* note 61 at 2.
[91] Id. at 9.
[92] Id. at 9-10.

Given the importance of semiconductors to the supply chain of consumer and military electronics, the industry has "[f]or decades" been considered relevant to national security within the US.[93] Concerns about the shift of semiconductor manufacturing capacity to the Asia-Pacific region have also dominated this discussion for decades, beginning with the rise of the Japanese semiconductor industry in the 1970s.[94] Worries about this transition has resulted in a range of different regulatory interventions in past decades to bolster and secure the US semiconductor industry, and block foreign access to the most cutting edge computational power.

Export controls in high-performance computing have been one important way these concerns have manifested as policy. Under Executive Order 13222, the US Commerce Department regulates exports of high performance computers (HPC) to certain countries, which are grouped into tiers.[95] Computers exceeding certain thresholds of processor performance to particular countries and end-users require prior government review for export.[96] These exports may be blocked on national security and anti-terrorism grounds.[97]

Fears around reliance on foreign manufacturing have also resulted in more extensive coordination between government and corporate actors. In 2004, the US Department of Defense (DOD) and the National Security Agency (NSA) initiated a "Trusted Foundry Program", which arranged for long-term contracts with accredited US companies to ensure that the government would have guaranteed access to trusted chips for its needs.[98] In practice, this program would center on a sole-source contract with IBM, which was deemed "the only U.S.-based company able to meet DOD and intelligence community needs for trusted leading-edge microelectronics."[99] While the program eventually expanded to include other trusted suppliers, one 2015 analysis by the Government Accountability Office (GAO) observed that IBM remained the only supplier with cutting-edge fabrication

---

[93] *Supra* note 61 at 21.
[94] Id. at 18-19. For a contemporaneous review, *see* Warren E. Davis & Daryl G. Hatano, *The American Semiconductor Industry and the Ascendancy of East Asia*, 27 California Management Review 128 (1985).
[95] Exec. Order No. 13,222, 66 Fed. Reg. 44025 (Aug 22, 2001); Bureau of Industry and Security, US Department of Commerce, Legal Authority for the Export Administration Regulations, https://www.bis.doc.gov/index.php/documents/Export%20Administration%20Regulations%20Training/876-legal-authority-for-the-export-administration-regulations/file.
[96] 15 CFR § 774, Supplement No. 1.
[97] Id.
[98] *Supra* note 61 at 22.
[99] Id.

facilities and that "use of accredited suppliers other than IBM has been minimal" as a result.[100]

This program is at present in transition. In 2014, IBM announced that it would transfer ownership of its foundries to the foreign owned GlobalFoundries.[101] This sale followed significant losses for IBM on its foundry business. During 2013 and 2014 the company lost $700 million on its two primary chip fabrication facilities, and in the end IBM paid $1.5 billion to GlobalFoundries for the acquisition.[102] This transfer was approved and included a multi-year contract with the DOD to provide semiconductors to the US government until 2023.[103] However, the 2015 GAO analysis concluded simply that "there are no near-term alternatives to the foundry services formerly provided by IBM."[104] As of late 2018, efforts are ongoing to identify "new approaches to retain trustable, leading-edge capabilities".[105]

Broader coordination has also happened in the past. During the 1980s and 1990s, concern around rising Japanese dominance in the space motivated the launch of SEMATECH, a public-private research consortium of US semiconductor firms.[106] Over subsequent years, $870 million in public subsidies from the Defense Advanced Research Projects Agency (DARPA) and augmented by matching funds from the participating companies focused on accelerating research into semiconductor manufacturing.[107] In 1996, after a period of growing US market share in semiconductors, the directors of SEMATECH voted to stop receiving federal funding.[108]

Many factors enabled the resurgence of the US semiconductor industry in the late 1980s and early 1990s, and the ultimate impact of SEMATECH has been disputed.[109] However, consensus favors the positive role the consortium played. One 2003 review by the National Academies concluded that the public-private effort was "key among elements in the industry's revival—contributing respectively to the restoration of financial health and product quality."[110]

---

[100] Marie A. Mak, Trusted Defense Microelectronics: Future Access and Capabilities Are Uncertain, Congressional Research Service 4 (2015), *available at* https://www.gao.gov/products/GAO-16-185T.
[101] *Supra* note 61 at 22.
[102] IBM-GlobalFoundries Deal Finalized | EE Times, EETimes, https://www.eetimes.com/document.asp?doc_id=1327029 (last visited Mar 18, 2018).
[103] Doug Cameron, Pentagon Hires Foreign Chips Supplier, Wall Street Journal, June 5, 2016, https://www.wsj.com/articles/pentagon-takes-foreign-chips-partner-1465159332.
[104] *See supra* note 95 at 4.
[105] Id.
[106] *See* Larry D. Browning & Judy C. Shetler, Sematech: Saving the U.S. Semiconductor Industry (2000).
[107] *See supra* note 22 at 19.
[108] Id. at 20.

Recent US Focus in the Space: China

Concerns around the robustness of the US semiconductor industry have persisted into the present as China has made a concerted push to advance its own semiconductor industry in recent years. In 2014, the State Council of China published *National Guidelines for Development and Promotion of the Integrated Circuit (IC) Industry,* a strategic plan which aims to establish national leadership across the semiconductor supply chain by 2030.[111] The plan allocates $100 to $150 billion from public and private investment to support this effort.[112] Goals declared later in 2015 would imply ambitions that "roughly all incremental foundry capacity installed globally over the next ten years would have to be in China.[113]

One result of this strategy has been a rapid expansion in acquisition activity by Chinese investment and technology companies. Some larger transactions in the space include a $2.3 billion acquisition of H3C, a Hong Kong subsidiary of Hewlett-Packard[114], a $2.75 billion acquisition of NXP Semiconductor's "Standard Products" division, and a $1.8 billion acquisition of Omnvision, which specializes in semiconductors for imaging applications.[115] One analysis in late 2017 estimated that "the total volume of transactions of China's semiconductor overseas M&As (completed) so far has exceeded US$11 billion."[116] It has also been reflected in a significant increase in the construction of new chip fabrication plants in the country.

---

[109] Id.
[110] National Academies, Government-Industry Partnerships for the Development of New Technologies 90 (2002), *available at* http://www.nap.edu/catalog/10584 (last visited Feb 15, 2018).
[111] Christopher Thomas, A new world under construction: China and semiconductors, McKinsey & Company, https://www.mckinsey.com/global-themes/asia-pacific/a-new-world-under-construction-china-and-semiconductors (last visited Mar 18, 2018).
[112] Id.
[113] Id.
[114] C. P. Yue & T. Lu, *China's Latest Overseas M&A in the Semiconductor Industry,* 9 IEEE Solid-State Circuits Magazine 8–12 (2017).
[115] Id.
[116] Id.

Roughly 40% of front-end semiconductor fabs slated to begin operation in 2017 to 2020 worldwide are located in China.[117]

This effort has also included the recruitment of several high-profile leaders within the Taiwanese semiconductor industry. This includes the recent move in 2017 by Shih-wei Sun, former CEO of Taiwan's UMC foundry, to China's Tsinghua Unigroup.[118] Similarly, Shang-Yi Chiang and Liang Mong-song, both research leaders at Taiwan's TSMC, were recruited to China's SMIC that same year.[119] These moves have produced some alarm within the Taiwanese semiconductor industry. TSMC chairperson Morris Chang has voiced concerns that the industry may become a "one way road" where "talent only departs and never arrives."[120]

These efforts are directed towards catching up: China has not traditionally been a leader in the semiconductor sector. While it accounts for a large percentage of worldwide consumption (57% in 2014), it largely relies on imports to meet this demand.[121] As of 2015, China possessed only around 6% of the most advanced semiconductor fabs globally.[122] One report noted that many fabrication plans continued to use "older technology and used equipment", which reflected China's focus on products that "do not require leading-edge semiconductors."[123] These recent efforts also follow on earlier, less successful initiatives by the Chinese government to accelerate the development of their national semiconductor base.[124] However, the present effort differs in its focus on fostering a defined set of national "champions" to compete internationally, rather than spreading financial support more thinly throughout the economy.[125]

---

[117] Dylan McGrath, China to house over 40% of semi fabs by 2020, EE Times Asia, https://www.eetasia.com/news/article/china-to-house-over-40-of-semi-fabs-by-2020 (last visited Mar 19, 2018).
[118] Cheng Ting-Fang, China poaches more Taiwanese chip talent, Nikkei Asian Review, https://asia.nikkei.com/magazine/20171109/Business/China-poaches-more-Taiwanese-chip-talent (last visited Feb 20, 2018); Alan Patterson, China Expected to Poach More Taiwan Chip Execs, EETimes, https://www.eetimes.com/document.asp?doc_id=1331144 (last visited Feb 20, 2018).
[119] Id.
[120] The China Post, TSMC boss slams government over China brain drain The China Post (2017), https://chinapost.nownews.com/20170125-11485 (last visited Mar 21, 2018).
[121] *See supra* note 61 at 14.
[122] Id. at 15.
[123] Dieter Ernst, *From Catching Up to Forging Ahead: China's Policies for Semiconductors*, East-West Institute (2015), https://www.ssrn.com/abstract=2744974 (last visited Mar 19, 2018).
[124] Gordon Orr & Christopher Thomas, Semiconductors in China: Brave new world or same old story?, McKinsey & Company, https://www.mckinsey.com/industries/semiconductors/our-insights/semiconductors-in-china-brave-new-world-or-same-old-story (last visited Mar 21, 2018).

The US has been active in attempting to counter these efforts. In 2016, then Secretary of Commerce Penny Pritzker declared that the US would "not allow any nation to dominate this [the semiconductor] industry and impede innovation through unfair trade practices and massive, non-market-based state intervention."[126] An expert committee convened by the NSA and the Department of Energy (DOE) that same year agreed, "expressing significant concern that – absent aggressive action by the U.S. – the U.S. will lose leadership and not control its own future in HPC [high-performance computing]."[127]

Several steps were taken towards this end. President Obama authorized the National Strategic Computing Initiative in 2015, which directed the DOD, DOE, and the National Science Foundation to work to "preserve [the US] leadership role in creating HPC technology".[128] In 2015, the administration restricted the sales of advanced microprocessors to several research sites associated with the Chinese supercomputer Tianhe-2.[129] The administration also blocked the acquisition of the US based assets of Aixtron, a German semiconductor manufacturer, by China's Fujian Grand Chip Investment Fund.[130] Similar deals were canceled on the threat of rejection by the Obama administration through the Committee on Foreign Investment in the United States (CFIUS), an inter-agency group "authorized to review transactions that could result in control of a U.S. business by a foreign person…in order to determine the effect of such transactions on the national security of the United States."[131] This included the sinking of a 2015 offer by China Resources Microelectronics and Hua Capital Management to acquire US-based Fairchild Semiconductor.[132]

---

[125] Chips on their shoulders, The Economist, Jan 23, 2016, https://www.economist.com/news/business/21688871-china-wants-become-superpower-semiconductors-and-plans-spend-colossal-sums (last visited Mar 21, 2018).
[126] U.S. Secretary of Commerce Penny Pritzker Delivers Major Policy Address on Semiconductors at Center for Strategic and International Studies, Department of Commerce (2016), https://www.commerce.gov/news/secretary-speeches/2016/11/us-secretary-commerce-penny-pritzker-delivers-major-policy-address (last visited Feb 20, 2018).
[127] U.S. Leadership in High Performance Computing (HPC), NSA-DOE Technical Meeting on High Performance Computing, https://www.nitrd.gov/nitrdgroups/images/b/b4/NSA_DOE_HPC_TechMeetingReport.pdf (last visited Feb 13, 2018).
[128] Office of Science and Technology Policy, Fact Sheet: National Strategic Computing Initiative (2015), https://obamawhitehouse.archives.gov/sites/default/files/microsites/ostp/nsci_fact_sheet.pdf.
[129] Don Clark, *U.S. Agencies Block Technology Exports for Supercomputer in China*, Wall Street Journal, April 9, 2015, http://www.wsj.com/articles/u-s-agencies-block-technology-exports-for-supercomputer-in-china-1428561987 (last visited Mar 21, 2018).
[130] Paul Mozur, *Obama Moves to Block Chinese Acquisition of a German Chip Maker*, The New York Times, December 2, 2016,

The Trump administration has maintained the active stance of the Obama administration in limiting foreign acquisition of US semiconductor companies. In 2017, CFIUS blocked a proposed acquisition of Lattice Semiconductor, a leading FPGA producer, by China Venture Capital Fund Corporation (CVCF), a state-run investment arm.[133] The decision cited "credible evidence that the foreign interest exercising control might take action that threatens to impair national security" and the "importance of semiconductor supply chain integrity to the U.S. government, and the use of Lattice products by the U.S. government."[134] In 2018, using the same authority and citing similar concerns, the Trump administration blocked a proposed $117 billion acquisition of Qualcomm by Singapore-based competitor Broadcom.[135] The administration similarly blocked a $580 million offer to acquire Xcerra, a Massachusetts-based semiconductor company, by state-backed investment funds Sino IC Capital and Hubei Xinyan that same year.[136] In late 2017, a bipartisan group proposed the

---

https://www.nytimes.com/2016/12/02/business/dealbook/china-aixtron-obama-cfius.html (last visited Mar 18, 2018).

[131] US Department of the Treasury, The Committee on Foreign Investment in the United States, https://www.treasury.gov/resource-center/international/Pages/Committee-on-Foreign-Investment-in-US.aspx (last accessed Mar 22, 2018).

[132] Diane Bartz and Liana B. Baker, Fairchild rejects Chinese offer on U.S. regulatory fears, Reuters, February 16, 2016, https://www.reuters.com/article/us-fairchild-semico-m-a/fairchild-says-china-resources-offer-not-superior-to-on-semis-idUSKCN0VP1O8 (last visited Mar 21, 2018).3

[133] Timothy B. Lee, Trump blocks Chinese purchase of US chipmaker over national security Ars Technica (2017), https://arstechnica.com/tech-policy/2017/09/trump-blocks-chinese-purchase-of-us-chipmaker-over-national-security/ (last visited Feb 13, 2018).

[134] Statement On The President's Decision Regarding Lattice Semiconductor Corporation, Department of the Treasury, https://www.treasury.gov/press-center/press-releases/Pages/sm0157.aspx (last visited Feb 13, 2018).

[135] Kate O'Keeffe, *Trump Orders Broadcom to Cease Attempt to Buy Qualcomm*, Wall Street Journal, March 13, 2018, https://www.wsj.com/articles/in-letter-cfius-suggests-it-may-soon-recommend-against-broadcom-bid-for-qualcomm-1520869867 (last visited Mar 21, 2018).

[136] Raymond Zhong, *U.S. Blocks a Chinese Deal Amid Rising Tensions Over Technology*, The New York Times, February 23, 2018,

Foreign Investment Risk Review Modernization Act (FIRRMA).[137] The bill would expand CFIUS review to include a range of transactions beyond outright acquisition, and would raise review in situations where an emerging technology prospectively "could be essential to national security."[138]

*Impact: Computational Power and the Geopolitics of Machine Learning*

From a geopolitical standpoint, the semiconductor industry finds itself in the middle of a perfect storm. First, semiconductor manufacturing capacity is considered a vital strategic asset, and the issue of who owns these businesses and their intellectual property is considered a national security matter. Second, the evolution of the industry has produced a bifurcated geography, with design-focused "fabless" foundries largely based in the US but with actual production taking place in the Asia-Pacific region. The result is that the semiconductor industry, like others, has been increasingly drawn into international politics as tensions escalate between the US and China.

This broader backdrop suggests that the more specific technology of machine learning is likely to be a flashpoint in the larger context of semiconductor competition between the US and China. In July 2017, the Chinese State Council released its "New Generation AI Development Plan", which sets out targets for development in AI and related industries. The plan declares AI to be "a new focus of international competition" and "a strategic technology that will lead in the future".[139] It plans for China to "firmly seize the strategic initiative in the new stage of international competition in AI development" and "effectively [protect] national security."[140] Specifically, the plan aims for China to be the world's "primary" leader in the technology and for the technology to be a $150 billion industry in the country by 2030.[141] As in the semiconductor context,

---

https://www.nytimes.com/2018/02/23/technology/china-microchips-cfius-xcerra.html (last visited Mar 21, 2018).
[137] Foreign Investment Risk Review Modernization Act of 2017, S. 2098, 115th Cong. (2017), https://www.congress.gov/bill/115th-congress/senate-bill/2098.
[138] Id.
[139] State Council of China, A Next Generation Artificial Intelligence Development Plan 2 (2017), *available at* https://www.newamerica.org/cybersecurity-initiative/blog/chinas-plan-lead-ai-purpose-prospects-and-problems/.
[140] Id.
[141] Id. at 6-7.

this plan echoes and builds on a series of earlier published policies focusing on robotics and other emerging technologies.[142]

There are a number of components to the State Council strategy. For one, the Chinese government intends to ramp up spending to support the development of AI in the country. Its Artificial Intelligence 2.0 program is poised to deploy "billions to develop AI for commercial and military use." [143] This will also be supported by a system of "government guidance funds" that provide financial support to companies and new ventures working on AI and related technologies.[144] The plan also contemplates recruiting and accelerating the training of leading AI researchers in the fields of "neural awareness, machine learning, automatic driving, intelligent robots, and other areas."[145]

Advancing computational power appears as a consistent theme in these plans. The State Council plan highlights the lack of "high-end chips" as a factor in China lagging in the field of AI.[146] The Ministry of Industry and Information Technology (MIIT), in elaborating on the State Council Plan, has explicitly laid out the "[development of] high-performance, scalable, and low-power cloud neural network chips for machine learning training applications" as a key priority.[147] It also distinguishes between the cloud and the separate need to advance chips to support "terminal" embedded applications, where "low-power, high-performance neural network chips suitable for machine learning algorithms" are needed.[148] By 2020, MIIT targets chips which will have "performance levels of 128 TFLOPS (16-bit floating point), and the energy efficiency ratio of more than 1 TFLOPS/Watt" with "terminal" chips having a similar efficiency.[149] Consistent with these policies, in August 2017 China's State Development & Investment Corporation led a $100 million round of funding in Cambricon, a startup focusing on the development of specialized machine learning hardware[150].

---

[142] *See, e.g.,* Central Committee of the Communist Party of China, The 13th Five-Year Plan for the Economic and Social Development of the People's Republic of China 67 (2016), *available at* http://en.ndrc.gov.cn/newsrelease/201612/P020161207645765233498.pdf (highlighting AI as a target "strategic emerging industry" for development); State Council of China, Guiding Opinions on Actively Promoting the "Internet Plus" Initiative (2015), *available at* http://www.gov.cn/zhengce/content/2015-07/04/content_10002.htm (highlighting the development of AI as a priority).
[143] Tom Simonite, The Trump Administration Can't Stop China From Becoming an AI Superpower, WIRED, Jun 29, 2018, https://www.wired.com/story/america-china-ai-ascension/ (last visited Mar 21, 2018).
[144] Yuan Yang, China fuels boom in domestic tech start-ups Financial Times (2017), https://www.ft.com/content/b63ee746-afc6-11e7-aab9-abaa44b1e130 (last visited Mar 21, 2018) (describing the government guidance funds ecosystem).
[145] *See supra* note 133 at 14.
[146] Id. at 4.
[147] Translation: Chinese government outlines AI ambitions through 2020, New America, https://www.newamerica.org/cybersecurity-initiative/blog/chinas-plan-lead-ai-purpose-prospects-and-problems/ (last visited Mar 21, 2018).
[148] Id.

National policies depend critically on what can be practically controlled by a state.[151] In a competitive environment, nations seek to control levers to shape the prospects of their adversaries, or at least the means of denying them access to key opportunities. Norms of open publication in research, the existence of widely distributed open-source toolkits, and the ever widening circle of those familiar with the practice of machine learning suggest that efforts to control research, software, or specialists in general will be challenging over time.[152] Computational power, with its small number of actors and large, fixed production facilities, is likely to be a focal point for national competition in the space.

International rivalry around the technology of machine learning will therefore in large part manifest itself in practice as international rivalry over the supply chain of computational power.

This is particularly true in the US-China context, where US companies depend on the manufacturing capability of a Chinese neighbor - Taiwan - and China desires access to more advanced computing power to support machine learning applications. Through market leaders Nvidia and Xilinx, Taiwanese foundries TSMC and UMC play a critical role in the manufacture of high-end GPUs and the FPGAs likely to support machine learning applications going forwards.

As CFIUS seems poised to make it increasingly challenging for China to pursue acquisitions of US semiconductor companies, shaping Taiwanese trade and foreign investment policies in semiconductors will assume greater importance.[153] The extent to which the US is able to successfully deny China access to advanced computing power, and the extent to which China is able to develop it domestically or acquire it otherwise, remains to be seen. The outcome will play a major role in determining the availability and sophistication of machine learning systems in different markets and among companies globally.

---

[149] Id.
[150] Tom Simonite, China Challenges Nvidia's Hold on Artificial Intelligence Chips | WIRED, https://www.wired.com/story/china-challenges-nvidias-hold-on-artificial-intelligence-chips/ (last visited Mar 19, 2018).
[151] Cf. James Scott, Seeing Like a State: How Certain Schemes to Improve the Human Condition Have Failed (New Ed edition ed. 1999).
[152] *See, e.g.,* arXiv, https://arxiv.org/ (open publication); Tensorflow, https://www.tensorflow.org/ (toolkit); Pytorch, http://pytorch.org/ (toolkit); Fast.ai, http://www.fast.ai/ (training); Deeplearning.ai, https://www.deeplearning.ai/ (training).

These trends may be further exacerbated as the machine learning field itself continues to evolve. In particular, there exist a rapidly developing set of research areas within machine learning - simulation learning, self-play, and meta-learning - which seem poised to heighten the importance of computational power as the key element in advancing the effectiveness of machine learning models. Moreover, meta-learning, which seeks to automate the architecture of machine learning algorithms, may also enable computational power to substitute somewhat for a lack of ready research and engineering specialists in the space.

**PART IV: COMPUTATIONAL POWER AS DATA SUBSTITUTE**

Computational power is just one ingredient in the process of training effective machine learning systems. Machine learning also requires sufficient and appropriate data, the examples on which the system will learn a good representation of the task it is attempting to accomplish.

The cost and availability of these inputs is crucial in thinking about the distributional impact of machine learning. Depending on the resources available to the practitioner and the presence of pre-existing data, acquiring both of these critical inputs may be more or less feasible for solving a particular problem. This influences both the set of actors that can generate novel machine learning systems, as well as the set of domains that machine learning can be effectively used in.

While computational power is increasingly commoditized and available as a service, acquiring and cleaning data appropriate to training a system to solve a given task remains stubbornly costly and time-consuming in a number of domains. Insofar as commercial actors continue to be the primary investors in the development of machine learning, this may render a set of potential machine learning applications infeasible or unprofitable to invest in developing.

---

[153] *See supra* note 113, Patterson (Quoting one industry analyst that "China will find it very tough to buy U.S. high-tech companies and difficult to leverage Chinese joint ventures or wholly-owned enterprises to access key U.S. intellectual property…We thus expect more senior Taiwan veterans to join China's semiconductor industry as a second wave of talent moves to China.")

However, one significant recent trend on this front has been a set of research successes which - in certain domains - are reducing a dependence on acquiring real-world data and increasing the role of computational power in the training of machine learning systems. This is significant since in these situations leveraging machine learning will be limited by the cost and availability of computing power, rather than the more fixed costs of gathering data in a specific domain. There are three arenas of research which seem particularly promising in this respect: simulation learning, self-play, and meta-learning.

Simulation Learning

Simulation learning seeks to replace the need for collecting real-world data by training an agent in a virtual space. For example, rather than training a robot arm to grasp objects by having a real robot arm attempting to pick up many different types of objects in the physical world, this approach trains the machine by having it "experience" picking up virtual objects in a computer simulation.[154] Once the system learns a good representation of the task, it can then be implemented to control a robot arm in the real world.[155] Successes in using this approach have prompted a number of different simulation environments for training agents to be released in recent years.[156]

The continuing advancement of simulation learning changes the balance of data and computational power. From a practical perspective,

---

[154] *Compare* Xue Bin Peng et al, Generalizing from Simulation, OpenAI Blog (2017), https://blog.openai.com/generalizing-from-simulation/ (last visited Mar 21, 2018) *with* Sergey Levine, Deep Learning for Robots: Learning from Large-Scale Interaction, Google Research Blog, https://research.googleblog.com/2016/03/deep-learning-for-robots-learning-from.html (last visited Mar 21, 2018).

[155] *See, e.g.,* Xue Bin Peng et al., *Sim-to-Real Transfer of Robotic Control with Dynamics Randomization*, (2017), https://arxiv.org/abs/1710.06537 (last visited Mar 21, 2018); Josh Tobin et al., *Domain Randomization for Transferring Deep Neural Networks from Simulation to the Real World*, arXiv:1703.06907 [cs] (2017), http://arxiv.org/abs/1703.06907 (last visited Mar 21, 2018).

[156] *See* Microsoft shares open source system for training drones, other gadgets to move safely on their own, Microsoft AI Blog (2017), https://blogs.microsoft.com/ai/microsoft-shares-open-source-system-training-drones-gadgets-move-safely/ (last visited Mar 21, 2018); DeepMind, DeepMind Lab, https://deepmind.com/research/publications/deepmind-lab/ (last visited Mar 21, 2018).

effective simulation learning removes the costs and logistical challenges of collecting data through sensors in the real world. In effect, it substitutes computational power for data as an input in machine learning. This is particularly promising since simulation learning might enable the collection of data far beyond what would be practical in the real world. Taking the example discussed above, a simulated approach may be able to run many more trials with virtual arms and virtual objects than the constraints of warehousing, maintenance, and energy in a real-world setting would allow. This at least theoretically holds out the promise of creating more effective machine learning systems which are trained on a more substantial simulated dataset at a faster rate.

These gains may be limited to particular types of domains. One obvious limitation of this approach is that it will only be effective in situations in which the simulation acts as a sufficiently realistic proxy for the actual challenge. A simulation which has unrealistic assumptions will cause the agent to learn a poor representation that fails when it is finally deployed "in the field."[157] Simulating a robot operating in the physical world, for instance, may be relatively more straightforward than simulating group social behavior or ecological systems where underlying forces are not well understood.

Self-Play

Simulation learning avoids the need for collecting real-world data by substituting a virtual environment for an agent to interact with. Self-play avoids the need for real-world data by having an agent play itself in order to improve and acquire "experience" in a given task. It therefore represents another method which effectively substitutes computational power for data as an input in the machine learning process.

At the time of writing, one notable demonstration of the use of self-play as a technique for reducing a dependence on data has been AlphaZero, a system designed by Google's DeepMind research lab which was able to achieve champion level skill at the board games of Go, chess, and shogi without the use of any pre-existing data of human games.[158] This marks a departure from earlier versions of these systems, which required training on data around historical examples of Go games to achieve a certain level of proficiency.[159] These self-play methods have also enabled advancements in transfer learning, allowing agents which learn skills in one environment to apply it in a new, novel one.[160]

---

[157] *See supra* note 148 at 1.
[158] David Silver et al., *Mastering Chess and Shogi by Self-Play with a General Reinforcement Learning Algorithm*, arXiv:1712.01815 [cs] (2017), http://arxiv.org/abs/1712.01815 (last visited Mar 21, 2018).

Like simulation learning, self-play is also somewhat domain limited to situations in which the competitive action of two agents will generate the data necessary to train the system to solve the desired problem. But, in these contexts, computational power becomes a primary rate limiter of progress, rather than the availability of data. As in the simulation learning case, sufficient computational power enables a multitude of agents to rapidly play a number of games which potentially outstrips the data that is available from games played by humans.

Meta-Learning

Learning architectures - the system by which a machine learning system extracts features and acquires a good representation of the task it is intended to solve - are another key ingredient in the machine learning workflow. For the purposes of this discussion, these architectures are notable since they influence the efficiency of the training process, defining how successful a trained machine learning system will be at a task for a given amount of training examples given to it.

These learning architectures have traditionally been hand-crafted by researchers and engineers developing these systems. Meta-learning seeks to automate this work by treating the task of designing these architectures as itself a machine learning task.[161] This approach has seen a number of promising successes, with machine generated learning architectures outpacing the benchmarked performance of their human crafted counterparts at the same task.[162] Meta-learning has also been used to optimize the distributions of neural networks across CPUs and GPUs and improve training and inference performance.[163]

---

[159] Id.
[160] *See, e.g.,* Trapit Bansal et al., *Emergent Complexity via Multi-Agent Competition*, arXiv:1710.03748 [cs] (2017), http://arxiv.org/abs/1710.03748 (last visited Mar 21, 2018).
[161] *See, e.g.,* Marcin Andrychowicz et al., *Learning to learn by gradient descent by gradient descent*, arXiv:1606.04474 [cs] (2016), http://arxiv.org/abs/1606.04474 (last visited Mar 21, 2018).
[162] *See, e.g.,* Esteban Real et al., *Regularized Evolution for Image Classifier Architecture Search*, arXiv:1802.01548 [cs] (2018), http://arxiv.org/abs/1802.01548 (last visited Mar 21, 2018).

Advancements in these techniques again have implications for the role of computational power in the workflow of machine learning. For one, meta-learning might improve the performance of machine learning systems with a given dataset, or enable equivalent performance with less data. Secondly, the automation of learning architecture design may also increase the speed at which new machine learning systems can be designed by aiding human technicians in creating the appropriate architecture for a given task.

Meta-learning does not offer a direct substitute to real-world data in the same way that simulation learning and self-play seem to. But, meta-learning is parallel to these methods insofar as it potentially reduces the amount of data needed to train a system to successfully accomplish a task. Meta-learning also accentuates the need for computational power, as a training process is required for generating an effective learning architecture as well as in learning how to accomplish the task itself.

*Impact: Computational Power and Economic Impact*

These shifts in the research field have broader significance because of the role that the inputs of machine learning play in defining its economic impact. Machine learning must be cost-effective and equivalently proficient at accomplishing a given task before it can serve as a competitive substitute for a human executing the same task. However, achieving this level of proficiency requires a machine learning model to be trained - a process that typically requires both sufficient computational power and sufficient data.

The availability of data and the cost of acquiring will vary depending on the domain. Existing market forces may already make this data freely available and structured for use. However, there are numerous reasons why this may not be the case. Legal restrictions rooted in copyright or privacy may make it expensive or challenging to acquire data.[164] The specific domain may itself make the practical tasks around acquiring the right data difficult. This is important because the poor availability of data can make it challenging for machine learning to produce a system which is an effective automated substitute for a given task. In the very least, these barriers may make the application of machine learning economically unviable for certain tasks and domains.

---

[163] Azalia Mirhoseini et al., *Device Placement Optimization with Reinforcement Learning*, arXiv:1706.04972 [cs] (2017), http://arxiv.org/abs/1706.04972 (last visited Feb 13, 2018).
[164] *See, e.g.,* Amanda Levendowski, How Copyright Law Can Fix Artificial Intelligence's Implicit Bias Problem (2017), https://papers.ssrn.com/abstract=3024938 (last visited Mar 21, 2018).

The advancements in machine learning research discussed above are significant because they reduce the dependence on data collection as a necessary component for training a machine learning system effectively. Simulation learning and self-play substitute computational power for data, enabling a system to learn a skill effectively without the physical costs and logistical requirements of real world collection. Meta-learning potentially boosts the efficiency of models by adding more computational power. These shifts have two economic impacts, one on the issue of technological unemployment, and the other on the issue of the industrial organization of machine learning.

Technological Unemployment

The economic literature examining the issue of technological unemployment in the context of machine learning has often abstracted away from the influence that inputs of data and compute have on the success of machine learning systems.[165] One method is to rely on surveys of expert opinion about the likelihood that a given task is broadly subject to automation by machines.[166] Common sources for these skills include the Dictionary of Occupational Titles (DOT) and its successor O*NET - databases compiled by the US Department of Labor.[167] These skills are usually articulated in a broad way, such as "[t]he ability to see objects in the presence of glare or bright lighting" or "[t]he ability to communicate information and ideas in speaking so others will understand."[168]

---

[165] *See, e.g.,* Anton Korinek and Joseph E Stiglitz, *Artificial Intelligence, Worker-Replacing Technological Change and Income Distribution*, NBER Economics of Artificial Intelligence 2017, http://www.nber.org/conf_papers/f100963/f100963.pdf; David Autor & Anna Salomons, *Robocalypse Now–Does Productivity Growth Threaten Employment?,* NBER Economics of Artificial Intelligence 2017, http://papers.nber.org/conf_papers/f100969/f100969.pdf.

[166] *See, e.g.,* Carl Benedikt Frey & Michael A. Osborne, *The future of employment: How susceptible are jobs to computerisation?*, 114 Technological Forecasting and Social Change 254–280 (2017).

[167] Dictionary of Occupational Titles, https://occupationalinfo.org/; O*Net Online, https://www.onetonline.org/.

[168] O*Net, Sensory Abilities, https://www.onetonline.org/find/descriptor/browse/Abilities/1.A.4/; O*Net, Cognitive Abilities, https://www.onetonline.org/find/descriptor/browse/Abilities/1.A.1/.

However, examining whether or not machine learning would enable the automation of "the handling of irregular objects" in such an abstract way may fail to take into account the costs and practicality of assembling the data necessary to train a system to an adequate level of proficiency in a domain. It may also mask challenges in training a system to be sufficiently robust to accomplish this task across a broad set of sectors and contexts. The data needed to train a machine learning model to grasp an irregular object in a well-lit, controlled environment may differ considerably from the training data needed to teach it to do so underwater, or in the dark. Examining the availability and cost of these data inputs therefore becomes critical since it can determine the practical terms under which a human skill is subject to automation.

The influence of data availability on machine learning performance suggests that the impact of machine learning on labor automation will proceed in a fragmentary, domain specific way. The data-replacing (or data-minimizing) techniques discussed above seem poised to accentuate this differential development. Though impressive, simulation learning and self-play are not all-purpose; they are most appropriate as learning schemes in specific circumstances. Self-play is specific to a reinforcement learning context with competing agents, and simulation learning is most effective in situations where it is feasible to construct a virtual space analogous to the real-world problem being solved. However, within these applicable domains, the barrier to machine learning being applied to a situation may be fixed by the cost of computing power, rather than the unpredictable costs and occasionally absolute barriers to collecting data on a particular task. That may accelerate displacement in those domains.

There are a scattered constellation of applicable domains suggested by these two techniques. Applications in robotics - in which an agent needs to accomplish a physical task in the real world, may be particularly benefitted by the acceleration made possible by simulation learning.[169] Competitive marketplaces, from capital markets to programmatic advertising, may be places where techniques such as self-play work to significantly improve machine learning system performance. Military autonomy may also be similarly impacted as self-play and simulation learning together allow the training of drones and other robotics in a manner it would otherwise be prohibitively expensive to do in the real-world. Within these types of contexts, machine learning may begin to have a disproportionate impact.

---

[169] *See, e.g.,* Cade Metz, *A.I. Researchers Leave Elon Musk Lab to Begin Robotics Start-Up*, The New York Times, November 6, 2017, https://www.nytimes.com/2017/11/06/technology/artificial-intelligence-start-up.html (last visited Mar 22, 2018).

Meta-learning impacts the relationship between machine learning, automation, and labor in a related, but more general way. Meta-learning may produce an overall improvement in the ability for systems to be trained to an adequate level of proficiency on a task with less data. This may make certain applications at the margin more viable than they would otherwise be, particularly in situations where data is scarce. These techniques may also lower the costs to creating new machine learning systems where engineering and development costs would have previously made investment in developing a capability unprofitable. Certain domains may have plentiful data, but the costs of developing and tuning an appropriate learning architecture and the potential profit of offering a product in that domain may render the choice to actually invest in developing the product unattractive. By aiding human domain experts in architecting these systems, meta-learning impacts the engineering time and cost of developing new systems and in doing so broadens the set of arenas that machine learning will be deployed in.

It is important to recognize that the processes by which labor is automated are complex, and not purely economic and technical. The ability for machine learning to proficiently accomplish a task is a necessary but not sufficient condition for that task to be actually automated in practice. Business practices can take a long time to respond to technological change.[170] Work is a sociotechnical system - and many forces shape who is employed and how work is accomplished. Automation may depend less on the economic opportunities presented by the technology, and more on a complex fabric of local historical and cultural factors[171]. Nevertheless, these techniques may open opportunities to automate where it was previously technically infeasible or cost-prohibitive to do so.

---

[170] James Manyika et al, A Future That Works: Automation, Employment, and Productivity, McKinsey Global Institute 70 (2017), *available at* https://www.mckinsey.com/mgi/overview/2017-in-review/automation-and-the-future-of-work/a-future-that-works-automation-employment-and-productivity (concluding that "automation will be a global force, but adoption will take decades and there is significant uncertainty on timing").

[171] *See, e.g.,* Alex Rosenblat and Tim Hwang, Regional Diversity in Autonomy and Work: A Case Study from Uber and Lyft Drivers (2016), *available at* https://datasociety.net/pubs/ia/Rosenblat-Hwang_Regional_Diversity-10-13.pdf (highlighting regional variation in machine learning driven workplaces); Marc Levinson, The Box: How the Shipping Container Made the World Smaller and the World Economy Bigger, Chapters 5-6 (2008) (discussing local political circumstances shaping industrial outcomes in shipping automation).

Industrial Organization

We can look at the economic impact of these research developments in two ways. Simulation learning, self-play, and meta-learning enable computational power to play a bigger role not just in defining the impact of machine learning on labor generally, but also in the competitive landscape between various firms seeking to offer products and services driven by AI, as well. To this end, computational power will play a role in shaping the industrial organization of machine learning: the types of competition, its level of consolidation, and the opportunities for new entrants.

Two "ingredients" serve as barriers to entry in machine learning: the data necessary to train the systems, and the relatively small pool of qualified talent necessary to build effective learning architectures.[172] Possessors of these assets have been able to generate some of the most impressive and groundbreaking machine learning systems, while competitors lacking these assets have been unable to compete on similar footing.[173]

These data-substituting and data-minimizing techniques are important because they shift the competitive balances between companies looking to lead in the latest generation of AI. For one, the advantages of data incumbency may be significantly eroded in certain markets. Expansive historical datasets on farm machinery to train agricultural robots, for instance, may be offset by sufficient computational power to simply simulate this farm machinery in a virtual space. As the research continues to develop, these may represent two viable paths towards developing and offering effective machine learning products in the space.

The benefits of these research developments might then redound to the handful of large technology companies that have been already investing most aggressively in the latest generation of machine learning such as Google and Facebook. These are companies which already possess significant endowments of computational power. On that count, the improvement of simulation learning or self-play may enable these existing technology leaders to enter new industries and provide products to meet use cases even where they lack training data. In the very least, it provides greater leverage to these technology companies in partnership negotiations with existing companies within a sector that do possess real world data.

---

[172] *See, e.g.,* Cade Metz, *Tech Giants Are Paying Huge Salaries for Scarce A.I. Talent*, The New York Times, October 22, 2017, https://www.nytimes.com/2017/10/22/technology/artificial-intelligence-experts-salaries.html (last visited Mar 21, 2018).

[173] *See, e.g.,* Tom Simonite, Apple's Privacy Pledge Complicates Its Push Into Artificial Intelligence, WIRED, Jul 14, 2017, https://www.wired.com/story/apple-ai-privacy/ (last visited Mar 21, 2018).

Meta-learning is likely to have a similar impact in practice, though in theory it should act as a leveler in the space. Meta-learning enables computational power to offset the advantages that a competitor may have in human machine learning expertise and talent. As the research matures and the techniques become increasingly commoditized, the autonomous construction of learning architectures may allow industries with extensive data and access to computing power to develop their own machine learning driven products effectively.

But, it is unclear at what rate meta-learning will become more commodified and usable by non-specialists. Though it seems poised to accelerate development times and improve the performance of learning architectures, meta-learning itself at the moment continues to require technical expertise and computational power to leverage effectively. Established technology companies may very well stand to gain again from these research developments, given their existing endowments of compute and expertise. In this view, meta-learning will tend to act as a force multiplier for entities which already have machine learning talent on hand, rather than democratizing the technology to entities with no experts at all.

These trends tend to suggest greater consolidation in the marketplace of companies offering products and services driven by machine learning. Simulation learning, self-play, and meta-learning all chip away at the potential advantage that non-AI focused industries might have in competing to offer their own machine learning systems: their data. In the very least, these trends suggest that one likely outcome is an ecosystem in which other industries rely on a handful of leading technology companies who provide these technologies as an "as a service" platform. This already seems to be in progress, as Google, Amazon, and a number of other companies compete to offer machine learning capabilities as a series of commodified APIs.[174]

**CONCLUSION: PROSPECTS FOR GOVERNANCE**

---

[174] *See, e.g.,* Google, Google Cloud: Machine Learning at Scale, https://cloud.google.com/products/machine-learning/; Amazon, Machine Learning on AWS, https://aws.amazon.com/machine-learning/.

Computational power does more than simply enable the current generation of breakthroughs in machine learning. Trends in computational architecture, semiconductor supply chains, and the field of machine learning itself suggest the multifaceted ways in which hardware plays a role in influencing the immediate and long-term impact of the technology on society.

These changes in the underlying hardware layer would appear to accentuate many existing concerns that have been raised around AI. Progression towards more specialized hardware seems to expand the privacy concerns presented by the technology, and present challenges to the nimble repair of these systems in the field. Evolving competition in the semiconductor industry seems likely to make machine learning a flashpoint in the broader economic conflicts between the US and China. Breakthroughs in simulation learning, self-play, and meta-learning seem likely to expand incumbent market power and potentially accelerate labor displacement.

Simultaneously, seeing these issues through the lens of hardware also raises opportunities for governance. Computational power is intimately linked to a nexus of issues in civil rights, consumer safety, geopolitical competition, labor policy, and corporate power. In contrast to the software and data that the practice of machine learning relies on, hardware and its production is more centralized, more able to be monitored, and less able to evade efforts to govern it. Hardware therefore offers one tangible point of control for those seeking to find levers on which to base policy and effectively shape the social impact of the technology.

Public policy to shape the markets and design of machine learning hardware might manifest in a range of different ways. Existing tools in the form of export controls and trade policy will play a significant role, especially given the specific political geography of the actors leading the design and actual fabrication of these chips. Industrial policy which aims to accelerate the development of a domestic semiconductor sector like SEMATECH, or otherwise ensure reliable access to computational power by companies and government actors working at the software layer of machine learning, will also play an important role. As FPGAs and ASICs broaden the potential range of contexts in which machine learning inference might be deployed, new regulations and industry standards might also serve to enshrine particular architectures for delivering machine learning products and services.

It is important to recognize that the current affordance that computational power provides to those thinking about the governance of AI may not be permanent. The machine learning field continues to evolve at a rapid rate, and new discoveries may change the relative importance and economics of computational power going forwards. Rapid advances in "one-shot learning", which aim to enable machines to train effectively on smaller numbers of examples; or "transfer learning", which works to allow machines to effectively use knowledge from one domain in another, may reduce the need for computational power. Indeed, a dependence on machine learning hardware and concerns about reliable access to it may itself motivate increased investment and more rapid progress on solving these types of technical problems.

As with simulation learning and self-play, it is possible that these research areas may have a narrower influence, influencing only specific domains and settings in which machine learning is deployed. Much of what we currently understand about machine learning suggests that significant computational power will continue to be a key component in building effective systems.[175] Regardless, any policy developed around such a fast-moving technology must take a potential future rewrite of our existing assumptions into account.

Multiple efforts have been launched to construct a global structure of governance around the research, development, and application of AI.[176] Many of these efforts are developing standards or sets of principles to guide the actions of the technology industry and the broader community of researchers and engineers working on machine learning.[177] The aim has been to attempt to shape norms around how the technology is applied, and the ethical responsibilities around the use of data as an input in the training process. This is one approach, one which is likely to become increasingly difficult to enforce effectively as the set of actors that might use the technology for good or for ill, responsibly or irresponsibly, continues to expand. As this transition occurs, governance efforts focused on shaping the direction and availability of computational power may offer a useful addition to the toolkit of policy options, and an increasingly important one at that.

---

[175] This is particularly the case given the increased performance that more data seems to promise. *See* Chen Sun et al., *Revisiting Unreasonable Effectiveness of Data in Deep Learning Era*, arXiv:1707.02968 [cs] (2017), http://arxiv.org/abs/1707.02968 (last visited Mar 21, 2018).

[176] *See, e.g.,* The Partnership on AI, http://partnershiponai.org/; The IEEE Global Initiative on Ethics of Autonomous and Intelligence Systems, https://ethicsinaction.ieee.org/; Building Agile Governance for AI & Robotics (BGI4AI), https://bgi4ai.org/

[177] *See, e.g.,* Partnership on AI, Tenets, https://www.partnershiponai.org/tenets/; IEEE, 7000 - Model Process for Addressing Ethical Concerns During System Design, http://standards.ieee.org/develop/project/7000.html; Future of Life, Asilomar AI Principles, https://futureoflife.org/ai-principles/; The Conference Towards AI Network Society, http://www.soumu.go.jp//000507517.pdf; DJ Patil, A Code of Ethics for Data Science (2018), https://medium.com/@dpatil/a-code-of-ethics-for-data-science-cda27d1fac1 (last visited Mar 22, 2018). For further sets of principles, *see, e.g.,* Accenture, An Ethical Framework for Responsible AI and Robots, https://www.accenture.com/gb-en/company-responsible-ai-robotics; The Future Computed: Artificial Intelligence and its role in society, The Official Microsoft Blog (2018),

---

https://blogs.microsoft.com/blog/2018/01/17/future-computed-artificial-intelligence-role-society/ (last visited Mar 21, 2018); Larry Dignan, IBM's Rometty lays out AI considerations, ethical principles ZDNet, http://www.zdnet.com/article/ibms-rometty-lays-out-ai-considerations-ethical-principles/ (last visited Mar 21, 2018).